# Joint Optimization and Variable Selection of High-dimensional Gaussian Processes

**Bo Chen** BCHEN3@CALTECH.EDU
California Institute of Technology, 1200 East California Boulevard, Pasadena, CA 91125, USA

**Rui M. Castro** RMCASTRO@TUE.NL
Eindhoven University of Technology, P.O. Box 513, 5600 MB Eindhoven, The Netherlands

**Andreas Krause** KRAUSEA@ETHZ.CH
ETH Zurich, Universitätstrasse 6, 8092 Zurich, Switzerland

## Abstract

Maximizing high-dimensional, non-convex functions through noisy observations is a notoriously hard problem, but one that arises in many applications. In this paper, we tackle this challenge by modeling the unknown function as a sample from a high-dimensional Gaussian process (GP) distribution. Assuming that the unknown function only depends on few relevant variables, we show that it is possible to perform joint variable selection and GP optimization. We provide strong performance guarantees for our algorithm, bounding the sample complexity of variable selection, and as well as providing cumulative regret bounds. We further provide empirical evidence on the effectiveness of our algorithm on several benchmark optimization problems.

## 1. Introduction

We consider the problem of function optimization in high dimensions. In many situations one wishes to find the maximum of a function quickly, from (noisy) evaluation at a small number of points. This problem occurs in various domains, for instance when learning optimal control strategies for robots (Lizotte et al., 2007), or when optimizing industrial processes that depend on many variables. It is particularly interesting to consider the case where the domain of the function $f$ we desire to optimize is high-dimensional (say $[-1,1]^D$), but when the values of the function depend only on a reduced, albeit unknown, set of variables. If there are $d$ such "active" variables, where $d \ll D$, it is somewhat plausible that the performance of such a function optimization procedure depends mostly on the intrinsic dimension $d$, and only depends mildly on the extrinsic dimension $D$. In this paper we formalize such insight, and provide a suite of algorithms based on Hierarchical Diagonal Sampling (HDS), which are able to perform both variable selection and function optimization in such settings. We provide strong theoretical guarantees, including a sample complexity bound that depends only logarithmically on the extrinsic dimensionality, as well as cumulative regret bounds on the performance of joint variable selection and optimization. We evaluate our proposed algorithms on several benchmark optimization problems.

**Related Work** Variable selection and optimization have both been extensively studied separately from each other. In variable selection one seeks an active set of variables, among many others, that explain a response function well. One family of models called sparse linear models study the case where the response function is *linear* in the variables. For example, Lasso (Tibshirani, 1996) tackles this combinatorial problem using an continuous approximation, which has been shown to be optimal under certain conditions (Donoho, 2006). Alternative models have been proposed to handle *non-linear* response functions. Automatic Relevance Determination (ARD, MacKay (1992)) is a Bayesian variable selection procedure that imposes a Gaussian prior on the bandwidths of the variables, which can be combined with a GP likelihood to handle non-linear functions. Yet there is little formal analysis regarding its sample complexity rate. Rodeo (Lafferty & Wasserman, 2008) is an efficient algorithm that simultaneously estimates bandwidths and selects variables in non-parametric regressions. It has favorable theoretical properties of risks and con-vergence rates. Also bearing similarities to our work is Distilled Sensing (Haupt et al., 2011), which attempts to quickly identify large portions of the variable space that are irrelevant, therefore reducing the search complexity as more data is collected, and effectively shedding the dependency on the extrinsic dimension. However, none of these models address the problem of function optimization in an active learning setting. The goal of active function optimization is to optimize an unknown function with as few samples as possible. One line of work called Bayesian global optimization (Ginsbourger & Riche, 2010; Brochu et al., 2010) assumes the unknown function is sampled from a GP. In particular, the GP-UCB (Srinivas et al., 2010) algorithm has been shown to have sub-linear regret and work well emprically. However, dealing with high dimensional domains is a notoriously hard challenge for these approaches. Most of this existing work has considered variable selection and function optimization separately. Recently, the problem of joint variable selection and linear optimization has been tackled by (Abbasi-Yadkori et al., 2012), who exploit sparsity to alleviate the curse of dimensionality. Concurrently, (Carpentier et al., 2012) combines compress sensing and bandit theory to achieve sub-linear regret bounds for sparse functions. However, they deal only with linear or approximately linear reward functions whereas our method handles non-linear functions.

## 2. Model and Problem Statement

We focus on functions of bounded domain, which, w.l.o.g., we assume to be $[-1,1]^D$, where $D$ is called the *extrinsic dimension*. Let $f : [-1,1]^D \to \mathbb{R}$ be a fixed, but unknown function. The value and location of the maximum of this function are our main quantities of interest. We assume that this function depends only on a subset of the domain variables, which we call the *active variables* or *active dimensions*, denoted by $\mathcal{A} \subset \{1,\ldots,D\}$. We are particularly interested in the case where the set of active dimensions is rather small relative to the extrinsic dimension, namely $d = |\mathcal{A}| \ll D$.

Without some regularity assumptions on $f$, optimization would be hopeless. We choose to model smoothness of $f$ by assuming that it is a sample from a Gaussian Process (GP, Rasmussen & Williams (2006)) with zero mean[1] and a squared-exponential[2] kernel $K$. In order to model the fact that the function depends on only a subset of the variables $\mathcal{A}$, we assume that the kernel is of the form $K(\mathbf{x},\mathbf{x}') = \sigma_s^2 \exp\Big(-\sum_{i\in\mathcal{A}} \frac{(x_i - x_i')^2}{b^2}\Big)$, where $\mathbf{x},\mathbf{x}' \in [-1,1]^D$, $\sigma_s^2$ is the self variance of the kernel, and $b > 0$ is the bandwidth corresponding to the active dimensions, respectively.

Although the function $f$ is modeled as a sample from a stochastic process, we assume it to be fixed throughout the data collection process. In particular we are allowed to collect data of the form $y_t = f(\mathbf{x}_t) + \epsilon_t$ where $t = \{1,2,\ldots\}$, $\epsilon_t$ are independent and identically distributed (i.i.d.) normal random variables with zero mean and variance $\sigma^2 > 0$ (assumed known), also independent of $f$. We are interested in developing an algorithm that chooses $\mathbf{x}_t$ as a function of all the past observations $\{\mathbf{x}_\ell, y_\ell\}_{\ell=1}^{t-1}$, in order to locate the maximum $\mathbf{x}^* = \operatorname{argmax}_{\mathbf{x}} f(\mathbf{x})$ of $f$ as quickly as possible. We evaluate any candidate algorithm in terms of its *regret*, $R_T = \sum_{t=1}^T [f(\mathbf{x}^*) - f(\mathbf{x}_t)]$. Notice that the average regret, $R_T/T$ is an upper bound on the minimum regret, $\min_{t=1,\ldots,T}[f(\mathbf{x}^*) - f(\mathbf{x}_t)]$, therefore minimizing the cumulative regret will lead to algorithms with good anytime performance.

## 3. Variable Selection

We propose a two-staged method for variable selection and function optimization, tied together through proper choice of certain parameters as described below. Variable selection is attained by means of a hierarchical diagonal sampling (HDS) stage. After the identification of active variables, we apply the GP-UCB algorithm (Srinivas et al., 2010) to optimize over the variables deemed active.

### 3.1. Hierarchical Diagonal Sampling

In a nutshell, the HDS algorithm recursively splits the set of variables into two sets of equal size, and keeps splitting the sets that are more likely to contain active variables. More specifically, HDS sequentially constructs a tree where each node corresponds to a set of variables, meaning each node can be uniquely identified by a subset of $\{1,\ldots,D\}$. Any node that is not a leaf has two children, corresponding to two disjoint subsets of dimensions, each with half of the size of the parent node. Each node in this tree can be in one of three states: **active** nodes contain at least one active dimension, **inactive** nodes are guaranteed (w.h.p.) to contain no active dimensions, and for **undetermined** nodes we have insufficient evidence to draw any conclusions about their activeness. All the nodes start undetermined, but as more samples are collected, a node will either become active, which implies that at least one of its children is active, or inactive, which renders its entire subtree inactive. Naturally, if a leaf node (which contains a single dimension) is active, then the dimension is deemed an active dimension.

[1] The assumption that the GP has zero mean is not critical, but greatly simplifies the description later. See Sec 2.7 of Rasmussen & Williams (2006) for a treatment of nonzero means.

[2] It is straightforward to extend our analysis to other isotropic kernel functions.

The crucial step in this algorithm is to determine if a node $I \subseteq \{1, \ldots, D\}$ is (in-)active. With that in mind, we construct a one-dimensional projection $f_I(\cdot)$ of the function defined as follows: Let $\mathbf{x}^{(0)} \in [-1,1]^D$ denote a randomly chosen *background vector* (this choice is made before any observations are collected). Define the function $\mathbf{x}_I : [-1,1] \to [-1,1]^D$ such that

$$\mathbf{x}_{I,i}(z) = \begin{cases} z & \text{if } i \in I \\ x_i^{(0)} & \text{otherwise} \end{cases} ,$$

where $z \in [-1,1]$. The one-dimensional projection of $f$ in $I$ is then simply defined as

$$f_I(z) = f(\mathbf{x}_I(z)) \ .$$

The function $f_I : [-1,1] \to \mathbb{R}$ is a sample from a one-dimensional GP, with kernel $K_I$

$$K_I(x,x') \quad = \sigma_s^2 \exp\left(-a_I \frac{(x-x')^2}{b^2}\right) \ ,$$

where $x, x' \in [-1,1]$ and $a_I = |\mathcal{A} \cap I|$ is the number of active variables in node $I$. In other words, $K_I$ is a squared-exponential kernel whose bandwidth depends on the number of active variables.

Therefore, to identify if $I$ contains any active variable it suffices to test which kernel best characterizes the landscape of noisy observations of $f_I$. We propose two methods for doing so: the Finite Difference Sequential Likelihood Ratio Test (FDT), and the GP Sequential Likelihood Ratio Test (GPT), each with their respective advantages.

### 3.2. Finite Difference Sequential Likelihood Ratio Test (FDT)

We use hypothesis testing in order to determine, whether node $I$ contains any active variables[3]. We consider two hypothesis: the null hypothesis $H_0$: $I$ contains no active variable; the alternative $H_1$: $I$ contains at least one active variable. We begin by considering a non-sequential testing approach to this problem.

**Finite Difference Testing:** The key idea is the following: If node $I$ contains no active variables, then $f_I(x)$ should be constant. In contrast, if the node $I$ contains active variables, $f_I(x)$ should exhibit a significant amount of variation as we vary $x$. In the following, we formalize this intuition. Suppose we pick two random points $x$ and $x'$, independently and uniformly distributed over $[-1,1]$. Consider $\Delta = \Delta(x,x') = f_I(x) - f_I(x')$. Both under $H_0$ and $H_1$ $\mathbb{E}[\Delta] = 0$. Further, in the null hypothesis, the variance $\mathbb{V}[\Delta] = 0$ as well. In contrast, under $H_1$, $\mathbb{V}[\Delta] = c > 0$. Unfortunately we cannot observe $\Delta(x,x')$ directly due to measurement noise. However, we can try to estimate $\mathbb{V}[\Delta]$ by picking $n$ pairs of points $x_i, x_i'$ independently at random and computing the test statistic

$$X_n = \frac{1}{2\sigma^2} \sum_{i=1}^{n} [y(x_i) - y(x_i')]^2,$$

where $y(x_i)$ and $y(x_i')$ are all independent noisy point observations of $f_I(\cdot)$, corrupted by additive Gaussian noise with zero mean and variance $\sigma^2$. Under $H_0$, $X_n$ is distributed according to a central $\chi_n^2$ distribution with $n$ degrees of freedom. In contrast, under $H_1$, $X_n$ is distributed according to a non-central $\chi^2_{n,B_n}$ distribution with (unknown) non-centrality parameter $B_n = \sum_i \Delta_i^2$, and $\Delta_i = (f_I(x_i) - f_I(x_i'))/\sqrt{2\sigma^2}$. The following Proposition provides a testing procedure, along with a sample-complexity bound, for distinguishing $H_0$ and $H_1$ with arbitrarily low failure probability.

**Proposition 3.1.** *Let $B_n = \sum_{i=1}^n \Delta_i^2$, where $\Delta_i^2 \in [0, M]$ are independent random variables satisfying $\mathbb{E}[\Delta_i^2] \geq c$. Consider testing between two hypothesis*

$$H_0 : X_n \sim \chi_n^2 \quad \text{and} \quad H_1 : X_n | B_n \sim \chi^2_{n,B_n} \ ,$$

*where in the alternative hypothesis we assume that, conditioned on $B_n$, the distribution of $X_n$ is a non-central $\chi^2$ with $n$ degrees of freedom and non-centrality parameter $B_n$. Provided*

$$n \geq \max\left\{2, \frac{16\left(1+\sqrt{1+c}\right)^2}{c^2}, \frac{2M^2}{c^2}\right\} \log(2/\alpha) \ .$$

*there is a thresholding test procedure that guarantees that both type I and type II error are less than $\alpha$. In other words, there is a value $\tau_n$ such that*

$$P_{H_0}(X_n > \tau_n) \leq \alpha \qquad \text{and} \qquad P_{H_1}(X_n < \tau_n) \leq \alpha \ .$$

Notice that the sample complexity given by Proposition 3.1 crucially depends on the lower bound $c$ on the variance, which can be viewed naturally as parameterizing the problem difficulty.

In order to apply this hypothesis testing strategy to our setting, we must ensure that samples from a GP satisfy $c = c(f) > 0$ with high probability over the random function $f$. We have the following result:

**Theorem 3.1.** *Let $\delta > 0$ and $\sigma_s^2 > 0$. Suppose $f$ is a sample from a GP on $[-1,1]$ with constant mean and covariance $k(x,x') = \sigma_s^2 \exp(-|x-x'|^2/h^2)$, for some $h \leq \frac{2}{(\log \frac{8}{\delta})^2}$. Let $x, x' \sim U([-1,1])$ be two independent, uniformly distributed random variables and define $\Delta = f(x) - f(x')$.*

[3] Using hypothesis testing for GP active learning was proposed by Krause & Guestrin (2007). However, their approach does not apply to our variable selection setting.

*There exist constants $a > 0$ and $b > 0$ such that with probability at least $1 - \delta$ over $f$ it holds for the conditional variance of $\Delta$ that*

$$\mathbb{V}_{x,x'}[\Delta \mid f] \geq \frac{\sigma_s^2 h^2}{4096 b^2 \log(2a/(h\delta))}.$$

Thus, as long as $h$ is sufficiently small, the variance $c$ of $\Delta$ is lower bounded with high probability. Asymptotically, if $h^{-1} = \Theta(\log(1/\delta)^2)$, then, as $\delta \to 0$,

$$\mathbb{V}_{x,x'}[\Delta \mid f] = \Omega\left(\frac{\sigma_s^2}{(\log(1/\delta))^4}\right),$$

with probability at least $1 - \delta$.

**A Sequential Testing Procedure:** While providing sample complexity guarantees, the bounds of Proposition 3.1 in conjunction with Theorem 3.1 are very loose in practice. Furthermore, in order to determine the threshold $\tau_n$, the lower bound $c$ on the variance must be taken in the worst-case scenario. As a more practical alternative, we consider a sequential testing strategy, which is able to adjust the sample complexity depending on the problem difficulty. The key idea behind our sequential approximation is that under the GP prior, we can characterize the distribution of $y_I(x) - y_I(x + \delta)$, the difference for point samples at a distance $\delta$.

First, we focus on the concrete case where node $I$ has either no active variables, or a known number of exactly $a$ active variables. In this case, the data distribution under each scenario is entirely known. The case $a = 1$ is the hardest, intuitively because $y_I$ varies less the fewer active variables $I$ has. Later we show that the composite case of distinguishing none vs. at least one active variable is of exactly the same difficulty. Setting $dy_I = y_I(x) - y_I(x + \delta)$, we have that the marginal distribution of this quantity is given by

$$dy_I \sim \mathcal{N}(0, \sigma_a^2); \quad \sigma_a^2 \quad \equiv 2\left[\left(1 - \exp\left(-a\frac{\delta^2}{b^2}\right)\right)\sigma_s^2 + \sigma^2\right].$$

Now suppose $a = 1$. If we pick $\delta$ at random, then $dy_I$ is distributed according to a scale-mixture of Gaussians. Instead, in our sequential test, we simply fix $\delta = 3b$. In this case, the variance under $H_0$ is $\sigma_0^2 \equiv 2\sigma^2$ whereas the variance under $H_1$ is $2((1 - 1/e^{-3})\sigma_s^2 + \sigma^2) \geq 2(.95\sigma_s^2 + \sigma^2) \equiv \sigma_1^2$. Thus by estimating the variance of the finite differences for this fixed choice of $\delta$, we expect to be able to distinguish between $H_0$ and $H_1$.

We now employ sequential hypothesis testing using the *sequential likelihood ratio test (SLRT)* as described in Siegmund (1985). This is an incremental procedure that sequentially computes the log likelihood ratio (LLR) between two hypotheses, and makes a decision once this ratio crosses two predetermined boundaries. In our finite differences setting, a pair of samples are collected each time to update the LLR between $H_1$ and $H_0$. The test terminates as soon as $LLR$ is either larger than an upper threshold $\Theta_1$ (and we accept $H_1$) or smaller than a lower threshold $\Theta_0$ (and we accept $H_0$). The LLR given a collection of $T$ samples $\{dy_I(x_t)\}_{t=1}^T$ can be computed in an additive fashion:

$$LLR(\{dy_I(x_t)\}_{t=1}^T) = \sum_{t=1}^{T} LLR(dy_I(x_t)), \qquad (1)$$

where the LLR for each individual $dy_I(x_t) = dy$ is:

$$\begin{aligned} LLR(dy) &= \log \mathcal{N}(dy|0, \sigma_1^2) - \log \mathcal{N}(dy|0, \sigma_0^2) \\ &= \left(\frac{1}{2\sigma_0^2} - \frac{1}{2\sigma_1^2}\right)dy^2 + \log\frac{\sigma_0}{\sigma_1}. \end{aligned} \qquad (2)$$

Several remarks are in order. First, for a fixed value of $dy$, under $H_1$, $\mathbb{E}[LLR(dy)]$ is a monotonic function in $\sigma_1$, which indicates that $I$ containing one active variable is indeed the hardest case. Secondly, the classical SLRT requires independent samples, and under the GP prior, two finite differences $dy_I(x) = y_I(x) - y_I(x + \delta)$ and $dy_I(x') = y_I(x') - y_I(x' + \delta)$ will be correlated. However, if $|x - x'|$ is sufficiently large, $dy_I(x)$ and $dy_I(x')$ will be nearly independent.

Lastly, since SLRT is carried out separately for each undetermined node in the tree, we would like to invest more samples on nodes that are the most likely to be active so as to reach all the active leaf nodes with minimum sample complexity. For this purpose, we allocate samples one at a time and always pick the node $I$ that has the largest $LLR$ so far. According to Eq. 2, statistics of $LLR$ per finite difference depend only on $\sigma_1$ and $\sigma_0$, hence all the nodes for which $H_1$ is true share the same upward slope for $LLR$, and the largest $LLR$ naturally is the most likely to reach $\Theta_1$.

### 3.3. GP Seq. Likelihood Ratio Test (GPT)

Instead of making an independence assumption, one can explicitly model the correlation between samples. Knowing the underlying hypothesis completely determines the data distribution, as it follows a GP with known covariance structure determined by $K_{I,a}$, where this kernel depends on $a$, the number of active variables in $I$. To avoid notational clutter, we drop the explicit dependence on node $I$ when its identity is clear from the context.

We focus first on a single node $I$. Given past observations of $\mathbf{x}_{1:t-1}$ and $\mathbf{y}_{1:t-1}$, we can compute the posterior distribution of $y_{I t}$ given $\mathbf{x}_t$ under each hypothesis.

$$y_{I t} \mid x_t, \mathbf{x}_{1:t-1}, \mathbf{y}_{1:t-1} \sim \mathcal{N}\left(\mu_a^t(x_t), (\sigma_a^t(x_t))^2\right).$$

$$\mu_a^t(x) \equiv \mathbf{k}_a(x)^T(\mathbf{K}_a + \sigma^2\mathbf{I})^{-1}\mathbf{y}_{1:t-1},$$
$$\sigma_a^t(x) \equiv \sigma^2 + K_a(x,x) - \mathbf{k}_a(x)^T(\mathbf{K}_a + \sigma^2\mathbf{I})^{-1}\mathbf{k}_a(x),$$
$$\mathbf{k}_a(x) \equiv [K_a(x,x_1), \cdots, K_a(x,x_{t-1})]^T,$$
$$\mathbf{K}_a \equiv [\mathbf{k}_a(x_1), \cdots, \mathbf{k}_a(x_{t-1})].$$

Consider the conditional $LLR_t(y \,|\, x_t, \mathbf{x}_{1:t-1}, \mathbf{y}_{1:t-1})$, denoted by $LLR_t(y)$ for convenience. Suppose we have sampled at $x_t^*$ and observed $y_t^*$. Then $LLR_t(y_t^*)$ is:

$$LLR_t(y_t^*) = \log \mathcal{N}(y_t^* | \mu_1^t(x_t^*), (\sigma_1^t(x_t^*))^2) - \log \mathcal{N}(y_t^* | \mu_0^t(x_t^*), (\sigma_0^t(x_t^*))^2).$$

Finally, the LLR of all the observed samples, $LLR_{1:t}$ is given by:

$$LLR_{1:t} = LLR_{1:t-1} + LLR_t(y_t^*). \quad (3)$$

After each observation we compare $LLR_{1:t}$ with two thresholds $\Theta_1$ and $\Theta_0$. If the LLR is above the first one, then we stop sampling and decide the node is active. If below the second threshold, we decide the node is inactive. Finally, if neither of these conditions holds we continue collecting data.

**Sampling Strategy:** The only remaining issue for this hypothesis testing procedure is to decide on the next sample. Note that, under $H_1$, the conditional distribution of the likelihood ratio for a sample at point $x_t$ follows a shifted non-central Chi-squared distribution:

$$LLR_t(y_t) \sim w_2\chi^2(1,\lambda) + w_0,$$

where $w_2 \equiv 0.5\left(\sigma_1^t(x_t)^2/\sigma_0^t(x_t)^2 - 1\right)$, $w_0 \equiv \log(\sigma_0^t(x_t)/\sigma_1^t(x_t)) - \frac{(\mu_1^t(x_t)-\mu_0^t(x_t))^2}{2(\sigma_1^t(x_t)^2-\sigma_0^t(x_t)^2)}$ and $\lambda = \left(\frac{\sigma_1^t(x_t)(\mu_1^t(x_t)-\mu_0^t(x_t))}{\sigma_1^t(x_t)^2-\sigma_0^t(x_t)^2}\right)^2$.

Given that we want to decide as quickly as we can if a node is active, it makes sense to choose the point $x_t$ that tends to maximize $LLR_t(y_t)$ the most. A natural choice is to take $x_t$ maximizing $\mathbb{E}(LLR_t(y_t)) + \sqrt{\mathbb{V}(LLR_t(y_t))}$. This choice is easily justified in light of tail bounds for Chi-squared distributions, such as those in the long version of the paper. Finally, as at each moment we are considering multiple nodes in the tree, we choose the node $I$ that maximizes the above quantity over all nodes. Let $UCB(I,x_t) = \mathbb{E}(LLR_{I,t}(y_t)) + \sqrt{\mathbb{V}(LLR_{I,t}(y_t))}$. We choose point $x_t$ and node $I$ so to maximize this index. In other words, we choose the node $I$ that is more likely to be deemed active after sampling. The complete procedure is described in Algorithm 1. Fig 1 shows an example of the search tree.

The following theorem characterizes the accuracy and sample complexity of HDS when using an arbitrary testing procedure block.

**Algorithm 1** Hierarchical Diagonal Sampling

**Input**: Sample budget; Thresholds $\Theta_1$, $\Theta_0$; $\mathcal{D}$.
Initialize root $I_0 \longleftarrow \mathcal{D}$, $LLR(I_0) \longleftarrow 0$
Initialize tree $\mathcal{T} \longleftarrow \{I_0\}$, active set $\mathcal{A} \longleftarrow \emptyset$.
**while** # samples $\leq$ budget and $|\mathcal{T}| > 0$ **do**
  **if** FDT **then**
    Sample node $I \longleftarrow \arg\max_{I' \in \mathcal{T}} LLR(I')$, then sample $x' \in [-1\ 1]$ uniformly at random.
  **else if** GPT **then**
    Sample $I, x \longleftarrow \arg\max_{I' \in \mathcal{T}, x' \in [-1\ 1]} UCB(I', x')$
    update index $UCB(I, x')\ \forall x' \in [-1\ 1]$
  Compute $LLR(I)$ using Eq 1 or 3
  **if** $LLR(I) \geq \Theta_1$ ($I$ is active) **then**
    **if** I is a singleton **then**
      $\mathcal{A} \longleftarrow \mathcal{A} \cup \{I\}$, $\mathcal{T} \longleftarrow \mathcal{T} \setminus \{I\}$
    **else**
      Split $I$ arbitrarily into two nodes $L$, $R$, each with half of the variables in $I$
      $LLR(L) \longleftarrow 0$, $LLR(R) \longleftarrow 0$.
      $\mathcal{T} \longleftarrow \mathcal{T} \setminus \{I\} \cup \{L, R\}$
  **else if** $LLR(I) \leq \Theta_0$ ($I$ is not active) **then**
    $\mathcal{T} \longleftarrow \mathcal{T} \setminus \{I\}$
  Increment # samples
**Output**: Set of active variables $\mathcal{A}$

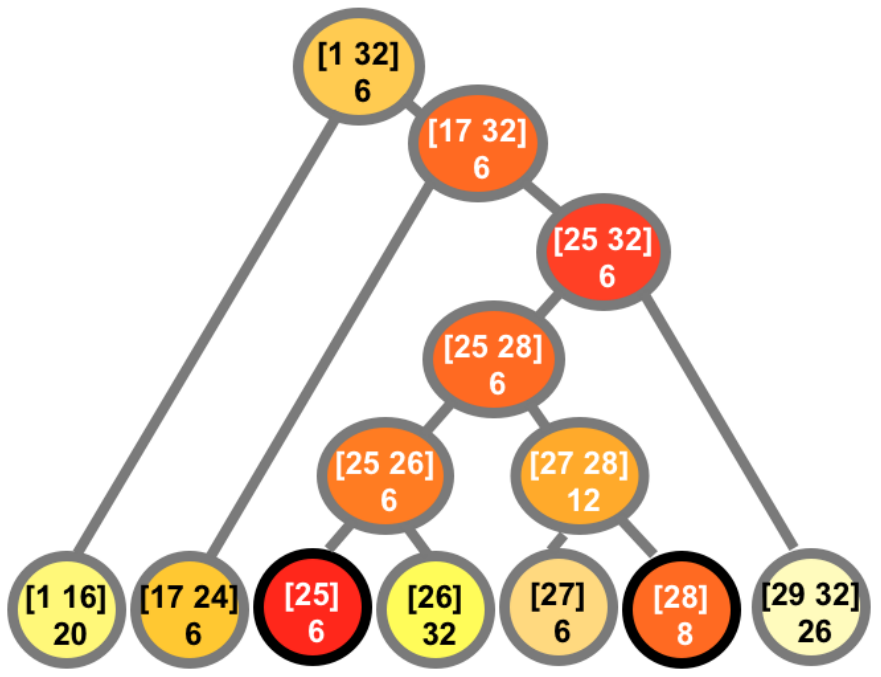

*Figure 1.* Sample HDS search tree. For each node, the set of dimensions is shown in brackets, followed by the number of samples. Each node is color-coded by the cumulative *LLR* over its samples, redder means larger. $\mathcal{A} = \{25, 28\}$.

**Theorem 3.2.** *Let $\mathcal{A}'$ denote the set of active dimensions identified by the* HDS *algorithm, also let $\alpha$ and $\beta$ denote respectively the false positive and false negative rates for the testing procedure used. All logs are base-2.*
***Accuracy:*** *For an arbitrary $\epsilon > 0$, provided $\alpha, \beta \leq (\epsilon/D)^{1/\lceil \log D \rceil}$ the probability of perfect recovery is $P(\mathcal{A}' = \mathcal{A}) \geq 1 - \epsilon$.*
***Sample Complexity:*** *Let $T_{max}$ be the maximum expected sample usage per node, and $N$ be the sample complexity of the* HDS *algorithm. The expected sample complexity is bounded by:*

$$\mathbb{E}(N) \leq \frac{2(1-\alpha)}{1-2\alpha} d \lceil \log D \rceil T_{max} \quad (4)$$

The proof of the first part expresses the probability of any particular leaf node being correctly classified,

and then applies a union bound on the event that all the leaf nodes are classified correctly. For the second part we study how testing errors change the tree generated by HDS under perfect conditions. Since every node incorrectly deemed active can spawn at most 2 inactive children with probability $\alpha$ we can get an upper bound on expected number of active nodes, and in turn, the worst case expected sample complexity. See the extended paper for details.

**Theorem 3.3.** *Given $\epsilon > 0$, set $\delta = \frac{\epsilon}{6d\lceil\log D\rceil}$ and $\alpha = (\epsilon/(2D))^{1/\lceil\log D\rceil}$. Assume $\epsilon$ is small enough so that $\alpha < 1/4$. Assume $b \leq \frac{2}{(\log(1/\delta)^2}$. Consider the HDS algorithm with non-sequential FDT using a fixed sample size per node $A_\epsilon \log(2/\alpha)$, where $A_\epsilon \equiv max\left\{2, \frac{16(1+\sqrt{1+B_\epsilon})^2}{B_\epsilon^2}, \frac{8}{B_\epsilon^2}\right\}$ and $B_\epsilon \equiv \sigma_s^2 b^2/(4096 c_2^2 \log(4c_1/(b\delta)))$, where $c_1, c_2 > 0$ are constants. Then, for a judicious choice of $c_1$ and $c_2$ the HDS-FDT procedure is correct with probability at least $1-\epsilon$, and the sample complexity:*

$$N < N_\epsilon \equiv A_\epsilon d\lceil\log D\rceil(\log 16 + \frac{\log(1/\epsilon)}{\lceil\log D\rceil}). \quad (5)$$

Note that $A_\epsilon = \Theta\left((\log(1/\epsilon) + \log(d\lceil\log D\rceil))^8\right)$, and $N_\epsilon = \Theta\left((\log\frac{1}{\epsilon} + \log(d\lceil\log D\rceil))^8 d(\lceil\log D\rceil + \log\frac{1}{\epsilon})\right)$. The proof (given in the extended paper) consists essentially in plugging in the results of Theorem 3.1 in Proposition 3.1, and applying theorem 3.2.

## 4. Optimization

After identifying the set $\mathcal{A}$ of active variables, we focus on optimizing $f$ over these relevant dimensions. In principle, various algorithms can be used for this purpose. We consider the GP-UCB algorithm (Srinivas et al., 2010). GP-UCB is a greedy algorithm, which iteratively picks the point

$$\mathbf{x}_{t+1} = \underset{x\in[-1,1]^D}{\text{argmax}}\ \mu_t(\mathbf{x}) + \beta_t^{1/2}\sigma_t(\mathbf{x}),$$

where $\mu_t(\mathbf{x})$ and $\sigma_t^2(\mathbf{x})$ are the posterior mean and variance at input $\mathbf{x}$, conditioned on the first samples $\mathbf{x}_1, \ldots, \mathbf{x}_t$ and associated observations $y_1, \ldots, y_t$. $\beta_1, \ldots, \beta_T$ is an appropriate sequence of constants for balancing exploration (choosing uncertain $\mathbf{x}$ with large variance) and exploitation (choosing $\mathbf{x}$ with large means), as specified in detail by Srinivas et al. (2010). For GP-UCB, strong performance guarantees are known: In particular, Theorem 2 of Srinivas et al. (2010) bounds the cumulative regret of GP-UCB in terms of the *maximum information gain* $\gamma_T$ obtainable by observing $f$ at an arbitrary set of $T$ inputs $\mathbf{x}_{1:T}$. Hereby, $\gamma_T$ is a monotonically increasing function of $T$, depending on the covariance function and domain of the GP. For the squared exponential kernel it is bounded by $O((\log T)^{d'+1})$, where $d'$ is the dimensionality of the underlying space. The cumulative regret of GP-UCB is bounded by $O^*(\sqrt{T\gamma_T})$ (where the $O^*$ notation hides logarithmic factors).

Straightforward application of GP-UCB without variable selection would lead to regret bounds depending on the extrinsic dimensionality $d' = D$. However, after variable selection we can apply GP-UCB only to variables that are deemed active, obtaining a regret bound depending on $d' = d$, the intrinsic dimensionality only.

Assume that function value $f(\mathbf{x})$ is bounded so that the maximum regret per sample is bounded by $C_0$. Let $N$ be the termination time of the HDS procedure, $\mathbf{x}^*$ a global optimum of $f$ and $R_T = \sum_{t=1}^T[f(\mathbf{x}^*) - f(\mathbf{x}_t)]$ the cumulative regret.

**Theorem 4.1.** *$\forall T \in \mathbb{N}^*, \delta_o > 0$, set $\epsilon = 1/T$ and*

$$\beta_t = 2\log(\frac{t^2 2\pi^2}{3\delta_o}) + 2d\log\left(\frac{2t^2 d\sigma_s}{b}\sqrt{\log(\frac{4d}{\delta_o})}\right).$$

*Running* HDS *with FDT and recovery rate $1-\epsilon$, followed by the* GP-UCB *algorithm on the variables deemed active guarantees that with probability $\geq 1-\delta_o$:*

$$\mathbb{E}(R_T) \leq A'_T\sqrt{T} + (N_\epsilon + 1)C_0 + 2 + \frac{N_\epsilon C_0}{T}.$$

*where $A'_T = \sqrt{C_1\beta_T\gamma_T} = O((\log T)^{d/2+1}$, $C_1 = 8/\log(1+b^{-2})$, and $N_\epsilon = O(\log(1/\epsilon)^9) = O((\log T)^9)$.*

Thus, the regret depends only logarithmically on the extrinsic dimension $D$.

The proof (given in the extended paper) bounds the worst case regret separately for the variable selection and optimization phases, assuming that maximum regret is incurred during HDS and, if HDS fails, during every round of the GP-UCB procedure. The former cases incurs linear regret for finite numbers of samples, and the latter does so for all samples but with a small probability $\epsilon$. When HDS is successful, Theorem 2 of Srinivas et al. (2010) guarantees a sub-linear regret bound for GP-UCB.

## 5. Experiments and Results

We compare HDS with a natural baseline called Coordinate-wise Sampling (CWS). CWS computes finite differences along each dimension separately using the same number of samples, and outputs the dimensions with the largest variance. We clairvoyantly choose the number of samples CWS needs to successfully recover all the active dimensions. Doing so favors the CWS algorithm in comparison to HDS.

**Functions sampled from a GP:** We consider the case where the test function is a sample from a GP with a squared-exponential kernel: $b = 0.1$

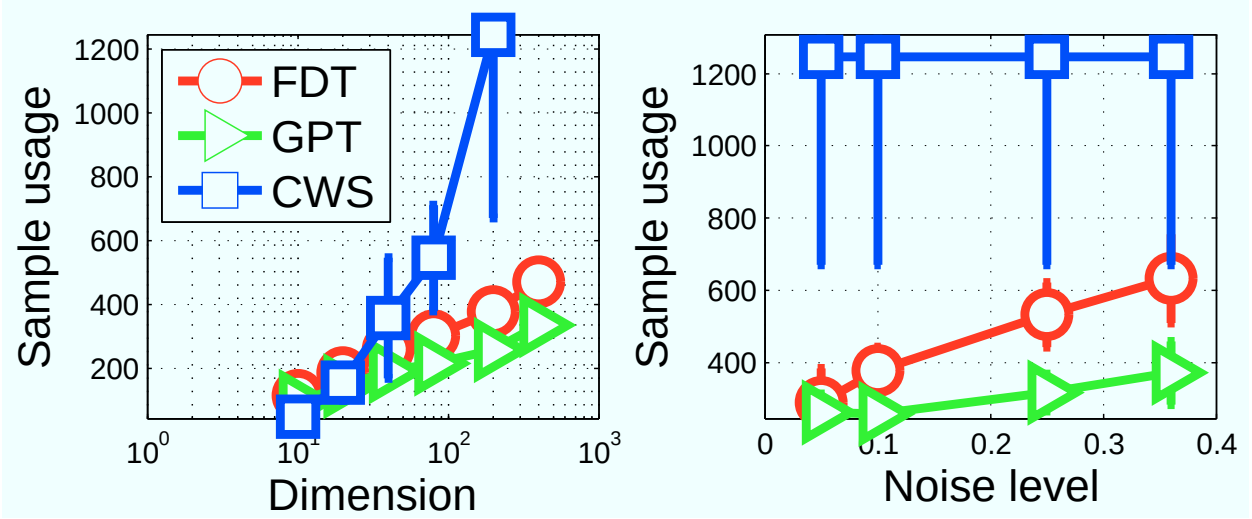

*Figure 2.* Sample complexity versus log dimensions and noise variance. If not varied, the default hyper-parameter setting is $D = 200$, and $\sigma_n^2 = 0.1$.

and self variance $\sigma_s^2 = 1$. We vary the total number of dimensions $D \in \{10, 20, 40, 80, 200, 400\}$, $\sigma_n^2 \in \{0.05, 0.1, 0.25, 0.36\}$, and compare FDT, GPT and CWS in terms of accuracy (recovery probability) and sample complexity. The thresholds $\{\Theta_1, \Theta_0\}$ were optimized using grid search over $\Theta_1 \in \{5, 10, 20\}$ and $\Theta_0 \in \{-5, -10, -20\}$ to ensure maximum recovery accuracy with the minimum sample complexity for the setting of $D = 200, \sigma_n^2 = 0.1$. The optimal value is then used for all settings. We repeat each setting for 20 random trials and report the mean $\pm$ 3 standard error. Accuracies of HDS under all settings are 100%. Fig 2 shows how sample complexity varies as a function of dimensionality $D$ and noise parameters $\sigma^2$. As predicted, the complexity grows linearly with $\log(D)$, the logarithm of the extrinsic dimension, and linearly with the noise level. GPT consistently uses about 50% less samples than FDT. In contrast, CWS is less stable and scales linearly as $D$. Using the oracle parameter setting CWS has little dependence on $\sigma^2$, yet HDS remains more efficient even in highly noisy situations.

We also examine the sensitivity of HDS w.r.t. the assumption that the noise parameter $\sigma^2$ and the bandwidth $b$ are known. Fig 3 shows accuracy and sample complexity as a function of the ratio between the misestimated $\sigma$ and the truth. FDT can tolerate over- and under-estimation of the noise level by a factor of 2, while GPT is more stable, withstanding misestimation by factors of 0.5 to 10. Both methods are robust w.r.t. $b$. Sample complexity and accuracy remain constant (FDT: 412 samples, 100% accuracy, GPT, 228 samples, 100% accuracy) when $b$ is misestimated by factors of 0.01 to 10.

**Functions embedded in high dimensions:** We also take the following low dimensional functions and hide them in a $D = 200$ dimensional space:

*Quad:* quadratic function ($d = \{2, 4, 6\}$).
$Quad(\mathbf{x}) = (P(\mathbf{x}-\mathbf{x}^*))^T(P(\mathbf{x}-\mathbf{x}^*))+\sigma_n^2$ where $\mathbf{x}^*$ is a random vector and $P$ is a diagonal projection matrix: $P_{ii} = 1/b$ if $i \in \mathcal{A}$ and $P_{ii} = 1/100$ otherwise.
*QuadMix:* quadratic with linear mixture ($d = \{2, 4, 6\}$). Similar to $Quad(\mathbf{x})$, $QuadMix(\mathbf{x}) = (MP(\mathbf{x}-\mathbf{x}^*))^T(MP(\mathbf{x}-\mathbf{x}^*)) + \sigma_n^2$ where $M = (1 - r_{mix})\mathbf{I} + r_{mix}$. $r_{mix} = 1/D$ controls how the irrelevant dimensions interfere with the relevant ones.
*Branin:* ($d = 2$) is a classical test function for unconstrained global optimization. It has a broad global landscape and peaks at $(-1, -1)$.
*Beale:* ($d = 2$) is a challenging test function. It remains flat for 90% in the domain, and gets flatter the closer it gets to the optimum.

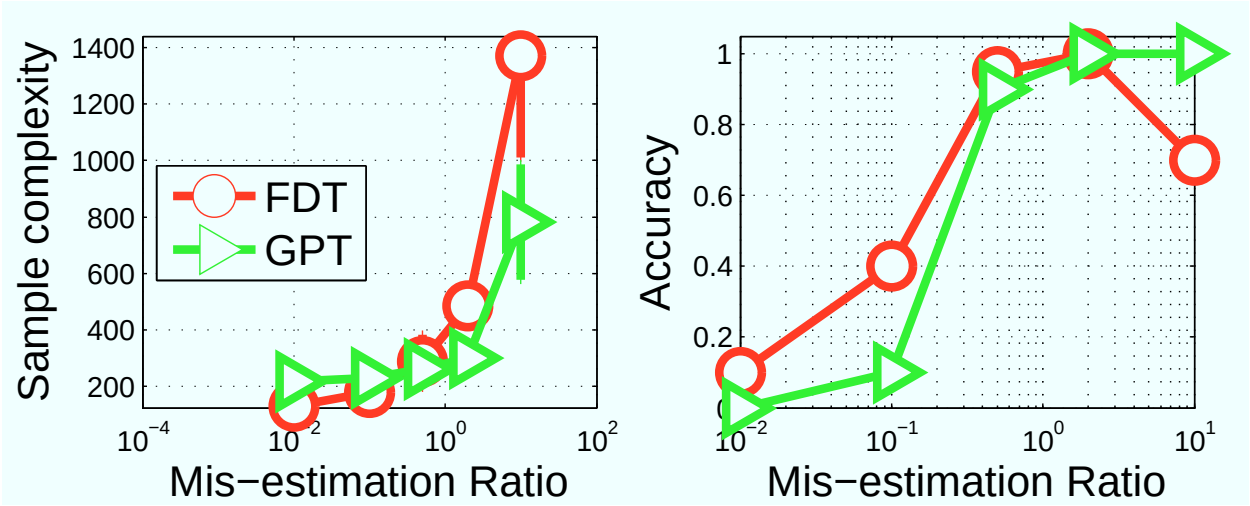

*Figure 3.* Sample complexity and accuracy versus misestimation ratio for noise level.

*Table 1.* Recovery Accuracy

| \Function | Quad | | | QuadMix | | | Branin | Beale |
|---|---|---|---|---|---|---|---|---|
| Method \ d | 2 | 4 | 6 | 2 | 4 | 6 | 2 | 2 |
| FDT | 1 | 1 | 1 | 1 | 1 | 1 | 1 | 0 |
| GPT | 1 | 1 | 1 | 1 | 1 | 0.95 | 1 | 0 |
| CWS | 0 | 0 | 0 | 0 | 0 | 0 | 0.45 | 0.25 |

We compare accuracy and sample complexity for the FDT, GPT and CWS. All functions are rescaled to $[-1\ 1]$ and the best parameter set found in the previous section is used[4]. The sample size is limited at 2000 per function. Error bars are obtained from 20 trials. The accuracies for Quad and QuadMix are shown in Table 1. The sample complexity for Quad and QuadMix are shown in Fig 4. The complexity for Branin is: FDT $(267 \pm 28)$, GPT$(236 \pm 12)$ and CWS $(1703 \pm 96)$. The complexity for Beale is: FDT$(280 \pm 10)$, GPT$(674 \pm 44)$ and CWS $(1802 \pm 82)$.

The results for Quad, QuadMix and Branin agree with the case of GP test functions and the theoretical analysis, showing the sample complexity's linear dependence on the relevant dimensions and logarithmic dependence on the extrinsic dimensionality. HDS (with testing blocks FDT or GPT) does not work for Beale because it is mostly constant, and therefore it is severely different than a typical sample from a GP.

**Joint Variable Selection and Optimization** Finally we compare the optimization performance of our 2-step procedure of HDS followed by GP-UCB against the conventional GP-UCB algorithm on all $D$ dimensions. Note that if $D$ is large, GP-UCB becomes in-feasible, in which case our method has a clear advantage. Even with small $D$, however, we show in Fig 5(b) and 5(c) that our method achieves a faster reduction in the average regret $R_T/T$, and obtains better minimum regrets $\min_t[f(\mathbf{x}^*) - f(\mathbf{x}_t)]$.

[4] In practice these parameters could be learned from a small set of held-out samples, or refined online.

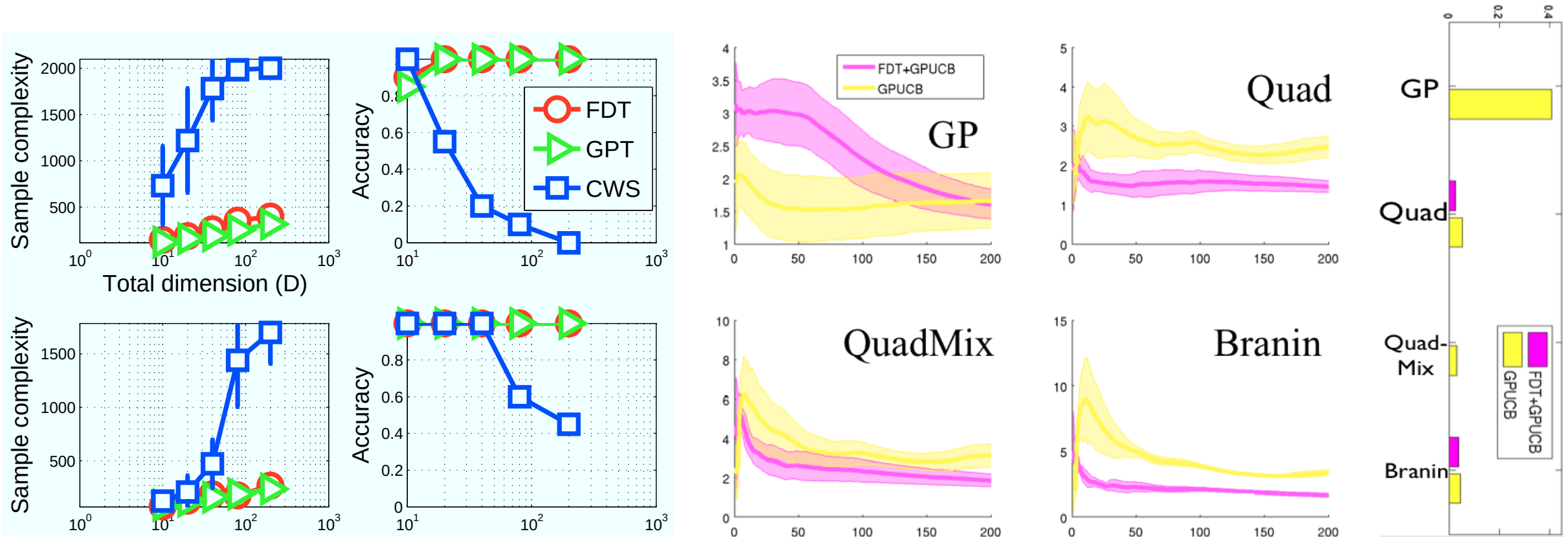

(a) *Sample complexity* (b) *Average Regret* (c) *Min Regret*

*Figure 5.* (a) Sample complexity and accuracy for Quad (upper) and Branin (lower) depending on $D$. (b) Average and (c) min regret of GP, Quad, QuadMix and Branin. $D = 4$, $d = 2$, $\sigma^2 = 0.1$. Confidence bands obtained from 20 trials.

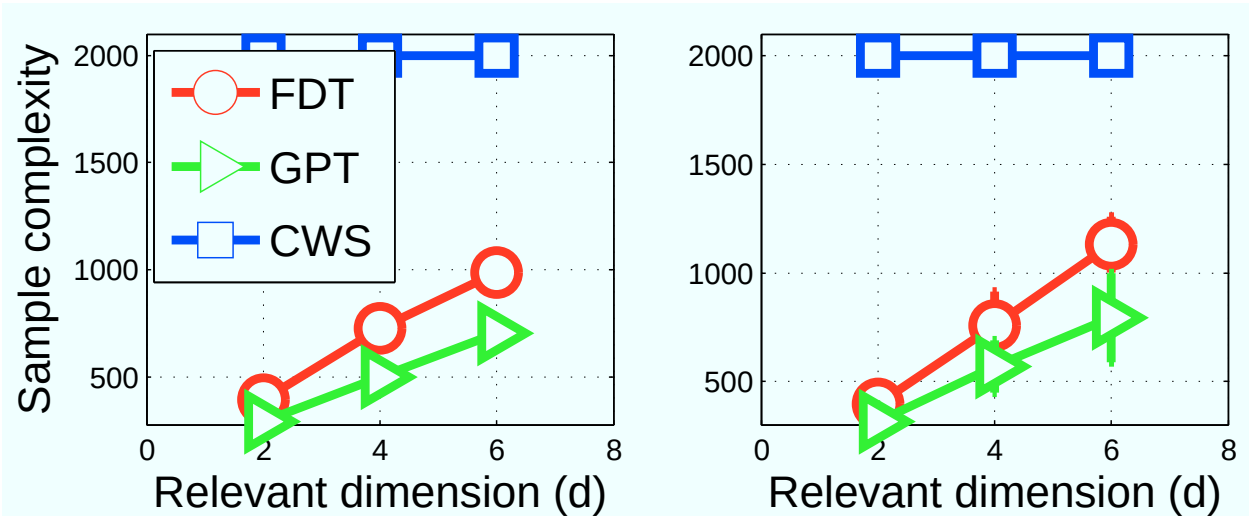

*Figure 4.* Sample complexity for Quad (left) and QuadMix functions(right).

## 6. Conclusions

We considered the problem of optimizing high dimensional functions that only depend on few active variables. We proposed HDS for variable selection and analyzed its sampling complexity in terms of properties of a modular hypothesis testing subroutine. For a classical (non-sequential) subroutine we proved sample complexity bounds, implying strong end-to-end performance guarantees for GP optimization in high dimensions. We also explored two practical alternatives based on sequential hypothesis testing and demonstrated their effectiveness on several high-dimensional optimization problems. We believe that our results provide important insights towards solving high dimensional optimization problems under uncertainty.

**Acknowledgments.** This work was partially supported by SNSF grant 200021_137971, NSF IIS−0953413 and DARPA MSEE FA8650-11-17156.